%% file: main.tex
\documentclass{article}

\usepackage[nonatbib]{neurips_2023}

\usepackage[utf8]{inputenc} % allow utf-8 input
\usepackage[T1]{fontenc}    % use 8-bit T1 fonts
\usepackage{hyperref}       % hyperlinks
\usepackage{url}            % simple URL typesetting
\usepackage{booktabs}       % professional-quality tables
\usepackage{amsfonts}       % blackboard math symbols
\usepackage{nicefrac}       % compact symbols for 1/2, etc.
\usepackage{microtype}      % microtypography
\usepackage{xcolor}         % colors
\usepackage{amssymb}% http://ctan.org/pkg/amssymb
\usepackage{pifont}% http://ctan.org/pkg/pifont
\usepackage{graphicx}
\usepackage{subfig}

\input{math_commands}

\newcommand{\cmark}{\ding{51}}%
\newcommand{\xmark}{\ding{55}}%

\title{Fine-Tuning the Retrieval Mechanism\\for Tabular Deep Learning}

\author{%
  Felix den Breejen \\
  KAIST AI\\
  \texttt{felixdenbreejen@kaist.ac.kr} \\
  \And
  Sangmin Bae \\
  KAIST AI \\
  \texttt{bsmn0223@kaist.ac.kr} \\
  \And
  Stephen Cha \\
  KAIST AI \\
  \texttt{jooncha@kaist.ac.kr\\} \\
  \And
  Tae-Young Kim \\
  KT \\
  \texttt{kim.taeyoung@kt.com} \\
  \And
  Seoung Hyun Koh \\
  KT \\
  \texttt{seounghyun.koh@kt.com} \\
  \And
  Se-Young Yun\dag \\
  KAIST AI \\
  \texttt{yunseyoung@kaist.ac.kr} \\
}

\nolinenumbers
\begin{document}

\maketitle

\begin{abstract}
% Tabular data finds widespread application in industry, yet deep learning techniques lag behind tree-based methods in this domain. Recent approaches incorporate a retrieval mechanism, which TabPFN utilizes for zero-shot prediction. Our study highlights the value of fine-tuning this method, demonstrating its superiority over standard neural network architectures. We emphasize the importance of pretraining and fine-tuning and investigate the role of the batch size. Our findings advocate for a future of tabular deep learning that leverages the retrieval mechanism for transfer learning instead of training from scratch.

While interests in tabular deep learning has significantly grown, conventional tree-based models still outperform deep learning methods. To narrow this performance gap, we explore the innovative retrieval mechanism, a methodology that allows neural networks to refer to other data points while making predictions.
Our experiments reveal that retrieval-based training, especially when fine-tuning the pretrained TabPFN model, notably surpasses existing methods. Moreover, the extensive pretraining plays a crucial role to enhance the performance of the model. These insights imply that blending the retrieval mechanism with pretraining and transfer learning schemes offers considerable potential for advancing the field of tabular deep learning.

\end{abstract}

\input{01-intro}

\input{02-related-work}

\input{03-methodology}

\input{04-Experiment}

\input{05-conclusion}

\clearpage

\section*{Acknowledgements}

This work was supported by Institute of Information \& communications Technology Planning \& Evaluation (IITP) grant funded by the Korea government(MSIT) 
(No.2019-0-00075, Artificial Intelligence Graduate School Program(KAIST)). This work was also supported by the “KT-KAIST Open R\&D” project funded by KT Corporation.

\bibliography{main}
\bibliographystyle{plain}

\appendix{\input{Appendix}}

\end{document}

%% file: math_commands.tex
\usepackage{amsmath,amsfonts,bm}

\def\eqref#1{equation~\ref{#1}}
\def\1{\bm{1}}

\def\vw{{\bm{w}}}
\def\vx{{\bm{x}}}
\def\vy{{\bm{y}}}
\def\vz{{\bm{z}}}

\def\mW{{\bm{W}}}
\def\mX{{\bm{X}}}

\DeclareMathAlphabet{\mathsfit}{\encodingdefault}{\sfdefault}{m}{sl}
\SetMathAlphabet{\mathsfit}{bold}{\encodingdefault}{\sfdefault}{bx}{n}

\def\gD{{\mathcal{D}}}

\def\sR{{\mathbb{R}}}

\def\sZ{{\mathbb{Z}}}

%% file: 01-intro.tex
\section{Introduction}

Tabular data has been widely utilized in the industry, leading to an increased interest in the research field of deep learning for tabular data. 
However, even with the significant advancements of deep learning in vision and language domains, previous studies \cite{grinsztajn_why_2022, mcelfresh_when_2023, shwartz-ziv_tabular_2022} have confirmed that tree-based methods, such as Random Forests \cite{breiman2001random} and XGBoost \cite{chen_xgboost_2016}, still prevail over deep learning techniques in tabular domains.
In order to narrow the performance gap, recent research has introduced architectures that leverage the decision tree algorithm \cite{popov2019neural, ke2019deepgbm} or the self-attention mechanism \cite{arik_tabnet_2021, huang_tabtransformer_2020, gorishniy2021revisiting, somepalli2021saint}. 
Moreover, several works have focused on transfer learning to exploit the benefits of deep neural networks by pretraining on extensive datasets \cite{yoon_vime_2020, bahri_scarf_2022, hollmann_tabpfn_2022, zhu2023xtab}.

Among various methodologies, a \textit{retrieval} mechanism, which retrieves data points as the reference for prediction, has opened up new avenues in the field of tabular deep learning \cite{kossen_self-attention_2021, somepalli2021saint, qin2021retrieval, hollmann_tabpfn_2022, gorishniy2023tabr}.
These neural networks make predictions by learning relationships between features and labels but also by referring to other samples, which resembles models like decision trees and nearest neighbors \cite{cover1967nearest}.
One specific application involves using input that contains a support set, $\gD = (\mX_\mathit{support}, \vy_\mathit{support}, \mX_\mathit{query})$, to predict the label of a query sample.

\begin{figure}[t!]
    \centering
    \vspace{-5pt}
    \includegraphics[width=\linewidth]{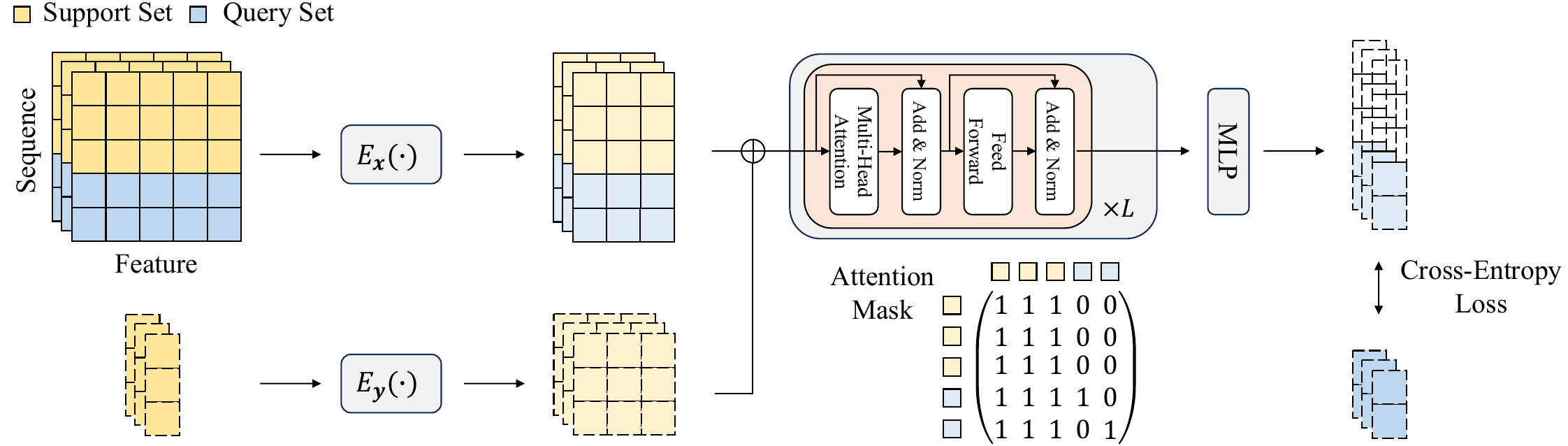}
    \vspace{-4pt}
    \caption{The overview of TabPFN training framework. Synthetically generated datasets are randomly divided into support and query sets during the pretraining phase, while train and test sets of tabular benchmarks are utilized during the evaluation.}
    \label{fig:tabpfn_overview}
    \vspace{-3pt}
\end{figure}

Specifically, TabPFN \cite{hollmann_tabpfn_2022} builds on prior-data fitted networks \cite{muller2021transformers} to pretrain on synthetically generated tabular prior datasets.
Their goal is to approximate the posterior predictive distribution of a test sample, which is conditioned on the set of training samples (see Figure~\ref{fig:tabpfn_overview}).
While the trained models exhibit strong performance on unseen real-world tabular datasets in a single forward pass, this in-context learning \cite{brown2020language} approach could have led to overlooking the further potential of retrieval-based training, potentially constraining the ability to transfer knowledge gained during pretraining.

In this paper, our comprehensive experiments, grounded on TabPFN and tabular benchmarks \cite{grinsztajn_why_2022}, reveal crucial insights emphasizing the significance of retrieval-based training. We demonstrate that the retrieval mechanism outperforms previous methods by further fine-tuning the pretrained TabPFN model. Moreover, a larger number of retrieved data points contributes to higher performance gains by offering enhanced references for prediction.
We also find that pretraining on extensive datasets plays an important role when comparing to models trained from scratch.
Consequently, we believe that integrating the retrieval mechanism with pretraining and transfer learning schemes will significantly influence the future landscape of tabular deep learning.

%% file: 02-related-work.tex
\section{Related Work}

\subsection{Retrieval Mechanism in Tabular Data}

Recent retrieval mechanism that implicitly or explicitly retrieves data points as the reference for prediction have shown significant improvements for tabular datasets.
Non-Parameteric Transformers \cite{kossen_self-attention_2021} take the entire dataset as input and learn relationships through self-attention between datapoints. 
Similarly, intersample attention from SAINT \cite{somepalli2021saint} associates the target row with other rows in the table.
On the other hand, there are explicit retrieval-based models, like RIM \cite{qin2021retrieval} and TabR \cite{gorishniy2023tabr}, that initially retrieve relevant instances using search engine techniques and a retrieval module, respectively.
Subsequently, they utilize the features and labels of the selected rows to make the final prediction for the target row.
Our work is primarily inspired by the TabPFN method \cite{hollmann_tabpfn_2022}, which let the token representations of test samples attend feature and label representations of training sets. While they only demonstrated the zero-shot evaluation performance, we fine-tuned pretrained models on actual tabular benchmarks, surpassing the previous deep learning approaches for tabular data.

% \subsection{Learning relations between samples}

% Recent approaches for tabular datasets have shown significant improvements by exploiting the relationships between samples.
% Non-Parameteric Transformers \cite{kossen_self-attention_2021} take the entire dataset as input and explicitly learn relationships through self-attention between datapoints. 
% Similarly, intersample attention, introduced by SAINT \cite{somepalli2021saint}, associates the target row with other rows in the table.
% On the other hand, there are retrieval-based models like RIM \cite{qin2021retrieval} and TabR \cite{gorishniy2023tabr} that initially retrieve relevant instances from the training data using search engine techniques and a retrieval module, respectively.
% Subsequently, they utilize the features and labels of the selected rows to make the final prediction of the target row.
% Our work is primarily inspired by the TabPFN method \cite{hollmann_tabpfn_2022}, which let the token representations of test samples attend the training set of feature and label representations. While they only demonstrated the zero-shot performance with a single forward pass, we fine-tuned pretrained models on actual tabular benchmarks, surpassing the previous deep learning approaches for tabular data.

\subsection{Tabular Transfer Learning} % TODO: revise with chatgpt and discussion
Leveraging transfer learning methods from computer vision \cite{chen2020simple} and natural language processing \cite{devlin2018bert} is a promising approach for tabular deep learning. Striving for generalization to unseen tables, various works in the tabular deep learning literature have attempted to pretrain on a large collection of different tables to exploit cross-table information \cite{zhu2023xtab, ye2023ct}, harness biases from LLMs \cite{nam2023semi, ye2023ct} and adopt self/semi-supervised learning \cite{nam2023stunt, nam2023semi}. XTab \cite{zhu2023xtab} processes entire tables using data featurizers and uses transformers for knowledge transfer. CT-BERT \cite{ye2023ct}, similar to XTab, employs a pretrained BERT for natural language and translates tables into "[column name] is [value]" strings. STUNT \cite{nam2023stunt} uses few-shot semi-supervised learning on self-generated tasks from unlabeled data, emphasizing column correlations. SPROUT \cite{nam2023semi} learns from unlabeled tables through LLMs, focusing on columns correlated with target labels, and applies in-context learning on generated prompts. The retrieval mechanism as used in this paper, can be seen as an alternative way of enabling transfer learning.

%Pretraining on a rich corpus of invariant features across tabular data enables high performance after fine-tuning. We remark that previous tabular deep learning pretraining approaches required a high volume of real tabular data, our fine-tuned TabPFN, which was pretrained solely on synthetic data, was able to outperform other NN-based methods. The implication would be that, in principle, our experiments could be compared to any sort of performance evaluation benchmark because there is no risk for testing on training data. 

%We note that the tabular deep learning literature lacks a standardized benchmark standard. Each work trains and evaluates on different data collections and splits. This makes fair comparison of the efficacy and efficiency of each method difficult. 

% 1) xtab, ct-bert, stunt, sprout 
% 2) address the lack of standardized benchmarks (pretrain set and evaluation set differences across works)

%% file: 03-methodology.tex
\section{Methodology}

TabPFN is a transformer-based architecture designed for zero-shot inference on small classification tasks. It is built on a basic transformer architecture that takes \((\mX_\mathit{support}, \vy_\mathit{support}, \mX_\mathit{query})\) as input. Each observation in the dataset is tokenized during the embedding process. For observation \(\vx_i \in \sR^{d_f}\) with corresponding class label \(y_i \in \sZ\) and \(d_f\) features, the embedding is calculated as follows:
\begin{align}
    \vz^\mathit{support}_i &= \mW_x \vx_i + \vw_y y_i \\ 
    \vz^\mathit{query}_i &= \mW_x \vx_i
\end{align}
% \begin{equation}
%     \vz^\mathit{support}_i = \mW_x \vx_i + \vw_y y_i,\quad \quad
%     \vz^\mathit{query}_i = \mW_x \vx_i
% \end{equation}

with weight matrices \(\mW_x \in \sR^{d \times d_f}\) and \(\vw_y \in \sR^{d \times 1}\) for hidden dimension \(d\).
These embedded tokens \(\vz_i\) are then processed through a basic transformer where attention is applied between the observations, with the exception of attention between pairs of inference observations. In the final layer, tokens corresponding to the inference observations are passed through a classification head. 
Figure~\ref{fig:tabpfn_overview} illustrates the overview of the TabPFN architecture.
% For an overview, see Figure \ref{fig:tabpfn_overview}

The features \(\vx_i\) are all variable-wise quantile transformed and scaled by \(\nicefrac{d_f}{d_f^i}\), where \(d_f^i\) is the effective number of features of observation \(i\). The target \(y_i\) is kept as an unnormalized integer representing the class index. Categorical features are not one-hot encoded before the quantile transformation. Under the pretraining settings outlined in the TabPFN paper, the model accomodates up to \(d_f=100\) features per observation and a maximum of 10 classes. % To account for the permutation invariance inherent in tabular data classification tasks, TabPFN's training procedure randomizes the order of features and classes in the input datasets.

% For details about the synthetic data generator, we refer to the TabPFN paper.

For fine-tuning TabPFN, we use the checkpoint from the pretrained TabPFN model. Given a real-life test dataset \( \gD = (\mX_\mathit{train}, \vy_\mathit{train}, \mX_\mathit{test}, \vy_\mathit{test})\), we create a distinct randomized 80\% split of \(\mX_\mathit{train}\) and \(\vy_\mathit{train}\) to make a new training dataset \( \gD_\mathit{train} = (\mX_\mathit{train}^\mathit{support}, \vy_\mathit{train}^\mathit{support}, \mX_\mathit{train}^\mathit{query}, \vy_\mathit{train}^\mathit{query}\)) at every training step. We fine-tune on these splits  and evaluate by predicting on \((\mX_\mathit{train}, \vy_\mathit{train}, \mX_\mathit{test})\).

% using fixed feature and class positions,

% <diagram of architecture>
% Explain in detail how tabPFN architecture and training procedure works

% Explain why this architecture is good for pretraining

% Show parallel with fewshot sentiment analysis on a new language pretrained on all languages.

% Discuss limitations of tabPFN

% Show how we finetune: hyperparameters, random positions -> fixed positions, training / validation split

% explain that TabPFN is zeroshot therefore prior matters a lot

%% file: 04-Experiment.tex
\section{Experiments}

We evaluate our fine-tuned TabPFN on tabular benchmarks \cite{grinsztajn_why_2022} consisting of approximately 50 datasets categorized based on feature type (numerical or numerical and categorical) and tasks (classification or regression). We focus on the medium-size benchmarks containing about 10,000 observations and less than a hundred features. To ensure consistency to benchmarks, we follow the same procedures for data splitting, optimizer selection, learning rate scheduling, early stopping, and accuracy or $R^2$ score normalization. 
% We exclude datasets with more than 100 features due to TabPFN limitations.

The hyperparameters are as follows. We use a learning rate of \(1.0 \times 10^{-5}\) and no weight decay for fine-tuning. 
% The variant without fine-tuning we refer to as "Zeroshot". 
When training from scratch, we use a learning rate of \(1.0 \times 10^{-4}\) and weight decay of \(1.0 \times 10^{-5}\). 
We use a support set size of 10,000 or 1,000. For a support set size of 10,000, the entire training dataset is processed through the transformer, while for 1,000, we randomly sample 1,000 samples from the training set and ensemble 10 times for each test observation. 
For all TabPFN variants, we run default settings without random search, as the search showed little benefit.  

\subsection{Main Results}

We compare our finetuned TabPFN with three tree-based methods (XGBoost \cite{chen_xgboost_2016}, Random Forest \cite{breiman2001random}, and GradientBoostingTree \cite{friedman_greedy_2001}) and four neural network-based methods (MLP, Resnet, SAINT \cite{somepalli2021saint}, and FT-Transformer \cite{gorishniy2021revisiting}). Since TabPFN pretraining relies on synthetically generated data, we consider this a fair comparison in the quantity of real-world data used.

% \begin{figure}[htp] 
% \centering
% \subfloat[Mixed features]{
%     \includegraphics[width=0.48\textwidth]{plots/mixed_classification_summary_BIG.png}    \label{fig:mixed_classification_summary}
%     }
% \hfill
% \subfloat[Numerical Features]{
%     \includegraphics[width=0.48\textwidth]{plots/numerical_classification_summary_BIG.png}\label{fig:numerical_classification_summary}
%     }
% \caption{Results of Fine-tuning TabPFN on classification tasks using the full 10,000 samples.}
% \label{fig:classification_summary}
% \end{figure}

\begin{figure}[!t]
    \centering
    \vspace{-3pt}
    \includegraphics[width=\textwidth]{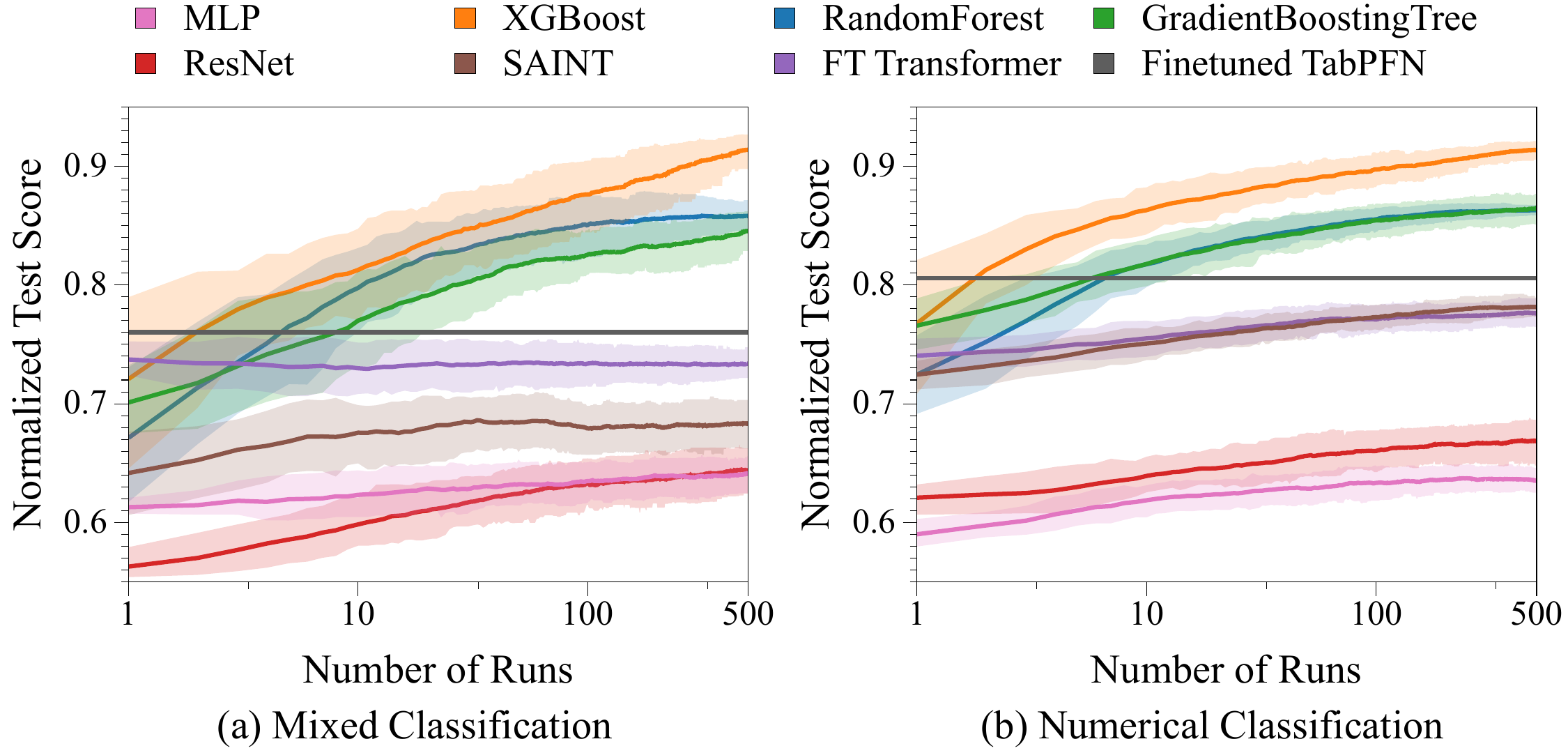}
    \caption{Comparison between fine-tuning TabPFN and baselines using the full 10,000 samples. }
    \vspace{-3pt}
    \label{fig:classification_summary}
\end{figure}

Figure \ref{fig:classification_summary} presents the overall comparison of classification accuracy. Here, we fine-tuned TabPFN with 10,000 number of retrieved samples. Utilizing default settings for fine-tuning TabPFN outperforms all neural network-based tabular data methods, even after extensive random hyperparameter searches for baselines. Furthermore, it also outperforms the tree-based methods under default settings, although tree-based methods tend to perform better on average after about 10 random searches. Additional dataset-specific results are available in Appendix \ref{sec:appendix_by_dataset}.

\subsection{Ablation Study}

% We also explore the importance of fine-tuning. 
Table~\ref{tab:tabpfn-variants} displays results for various TabPFN variations on mixed and numerical classification tasks, referring to the settings in Figure~\ref{fig:classification_summary}. There is a significant gap in normalized accuracy between fine-tuning on 10k samples and learning from scratch, with Zeroshot performance falling in between. We believe that these results imply the importance of fine-tuning phase after pretraining.

% \begin{table}
% \caption{Results on TabPFN variants for classification accuracy and regression $R^2$.}
% \label{tab:tabpfn-variants}
% \begin{tabular}{l|p{2cm}p{2cm}p{2cm}p{2cm}}
% \toprule
%  & Mixed \newline Classification & Numerical \newline Classification & Mixed \newline Regression & Numerical \newline Regression \\
% \midrule
% TabPFN Finetune 1k & 0.6221 & 0.7073 & 0.0677 & 0.0877 \\
% TabPFN Finetune 10k & \textbf{0.7598} & \textbf{0.8057} & 0.0353 & 0.0913 \\
% TabPFN Scratch & 0.5008 & 0.6057 & \textbf{0.2199} & \textbf{0.2906} \\
% TabPFN Zeroshot 1k & 0.4451 & 0.5421 & 0.0000 & 0.0000 \\
% TabPFN Zeroshot 10k & 0.5989 & 0.6609 & 0.0000 & 0.0000 \\
% \bottomrule
% \end{tabular}
% \end{table}

\begin{table}
\caption{Results on TabPFN variants for classification accuracy and regression $R^2$ score. The pretraining and fine-tuning are conducted on synthetic datasets and actual tabular datasets, respectively. ICL denotes the in-context learning, Scratch denotes training from scratch.}
\label{tab:tabpfn-variants}
\begin{tabular}{lccc|cccc}
\toprule
 &  &  &  & \multicolumn{2}{c}{classification} & \multicolumn{2}{c}{regression} \\
\cmidrule(l{2pt}r{2pt}){5-6} \cmidrule(l{2pt}r{2pt}){7-8}
method & pretrain & fine-tune & \# retrieval & mixed & numerical & mixed & numerical \\
\midrule
Scratch & \xmark & \xmark & 10k & 0.5008 & 0.6057 & \textbf{0.2199} & \textbf{0.2906} \\
ICL & \cmark & \xmark & 1k & 0.4451 & 0.5421 & 0.0000 & 0.0000 \\
ICL & \cmark & \xmark & 10k & 0.5989 & 0.6609 & 0.0000 & 0.0000 \\
Fine-tune & \cmark & \cmark & 1k & 0.6221 & 0.7073 & 0.0677 & 0.0877 \\
Fine-tune & \cmark & \cmark & 10k & \textbf{0.7598} & \textbf{0.8057} & 0.0353 & 0.0913 \\
\bottomrule
\end{tabular}
\end{table}
We also examine the size of the support set. TabPFN is pretrained on datasets smaller than 1,000 samples. The classification ICL and Fine-tuning results indicate that processing 10,000 samples is preferable to ensembling 1,000-sample segments. For large datasets with over one million samples, the best approach appears to be utilizing as many observations as the GPU memory allows.

In regression tasks, fine-tuned TabPFN underperforms, as the expected \(R^2\) should be around 0.8 (see Appendix \ref{sec:appendix_regression_summary}). This results comes as no surprise, as TabPFN is not pretrained for regression. The model's embedding expects integer class labels, but we provide quantile-transformed regression targets. This suggests the need for pretraining on regression or exploring alternative methods.

% - Performance is especially good on datasets where tabular NN methods struggle.

%% file: 05-conclusion.tex
\section{Conclusion}

Our findings suggest that fine-tuning tabular data transformers pretrained using a retrieval mechanism is a compelling avenue for improving neural network based tabular data prediction. While research was predominantly centered on training neural networks from scratch, our results suggest that adopting the transfer learning paradigm holds significant potential. Pretrained transformers, unlike small tabular data architectures, can scale effectively in the size of the network. We imagine that large-scale companies could take this concept by developing a billion parameter transformer fine-tunable on downstream datasets. Such an approach, we believe, has the potential to decisively surpass the performance of all tree-based methods.

Our analysis identifies several areas for future research, with one pressing concern the scalability in the number of observations. Models like GPT-4 \cite{openai2023gpt4} are limited by a context length of 32k tokens and CodeLlama \cite{rozière2023code} by context length of 100k, but real-world tabular datasets can have millions of observations. Further research areas include refining the architecture to tailor it specifically to tabular data, exploring the limitations of synthetic data, and enabling regression within the model. In essence, our study opens doors to a range of exciting possibilities for enhancing tabular data prediction through transfer learning, paving the way for general tabular data solver.

%% file: Appendix.tex
\clearpage

\section{Normalized Graphs for Regression from Scratch}\label{sec:appendix_regression_summary}

\begin{figure}[hbt]
\centering
\includegraphics[width=0.7\textwidth]{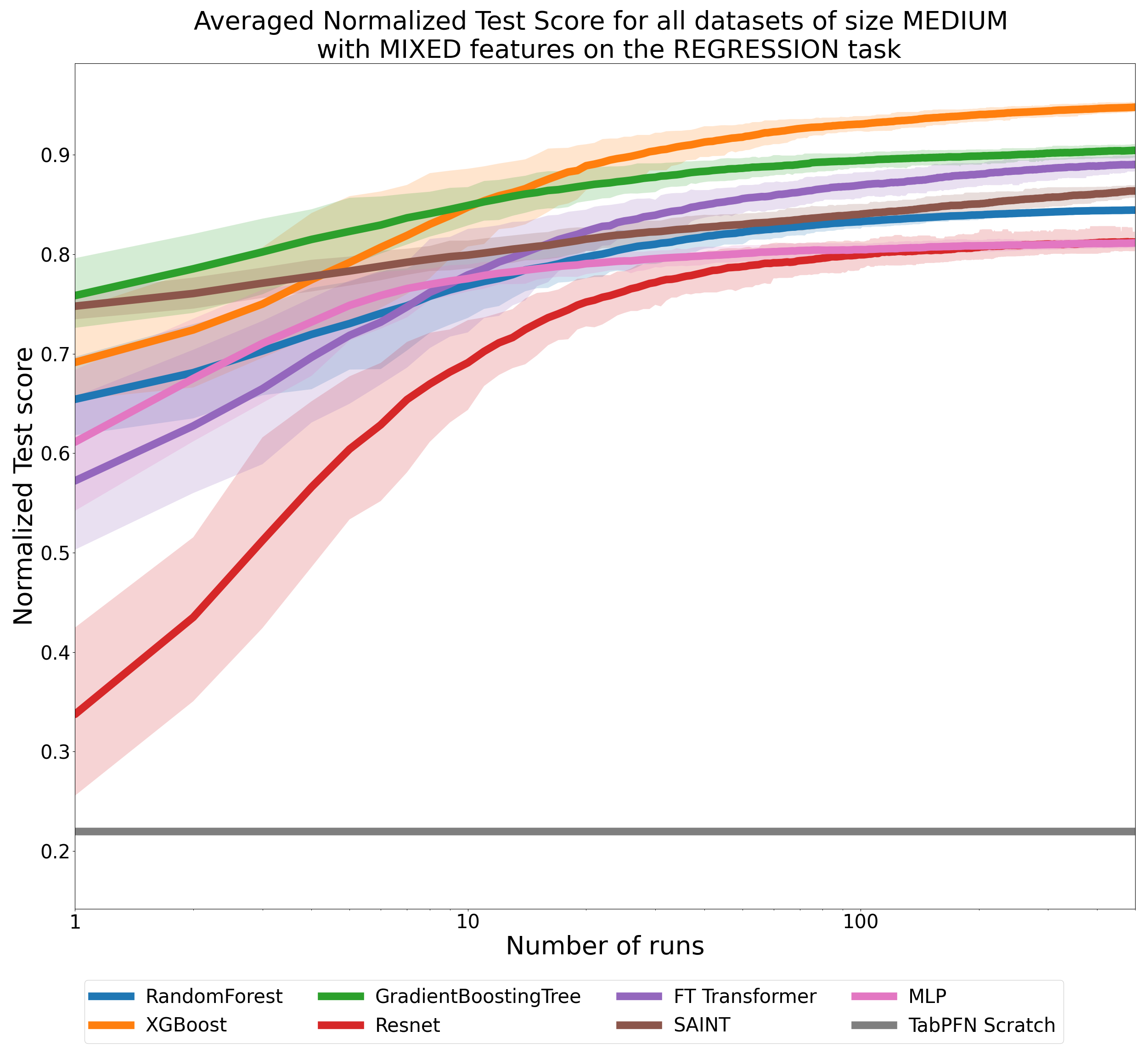}%
\label{fig:numerical_regression_summary}
\end{figure}

\begin{figure}[hbt]
\centering
\includegraphics[width=0.7\textwidth]{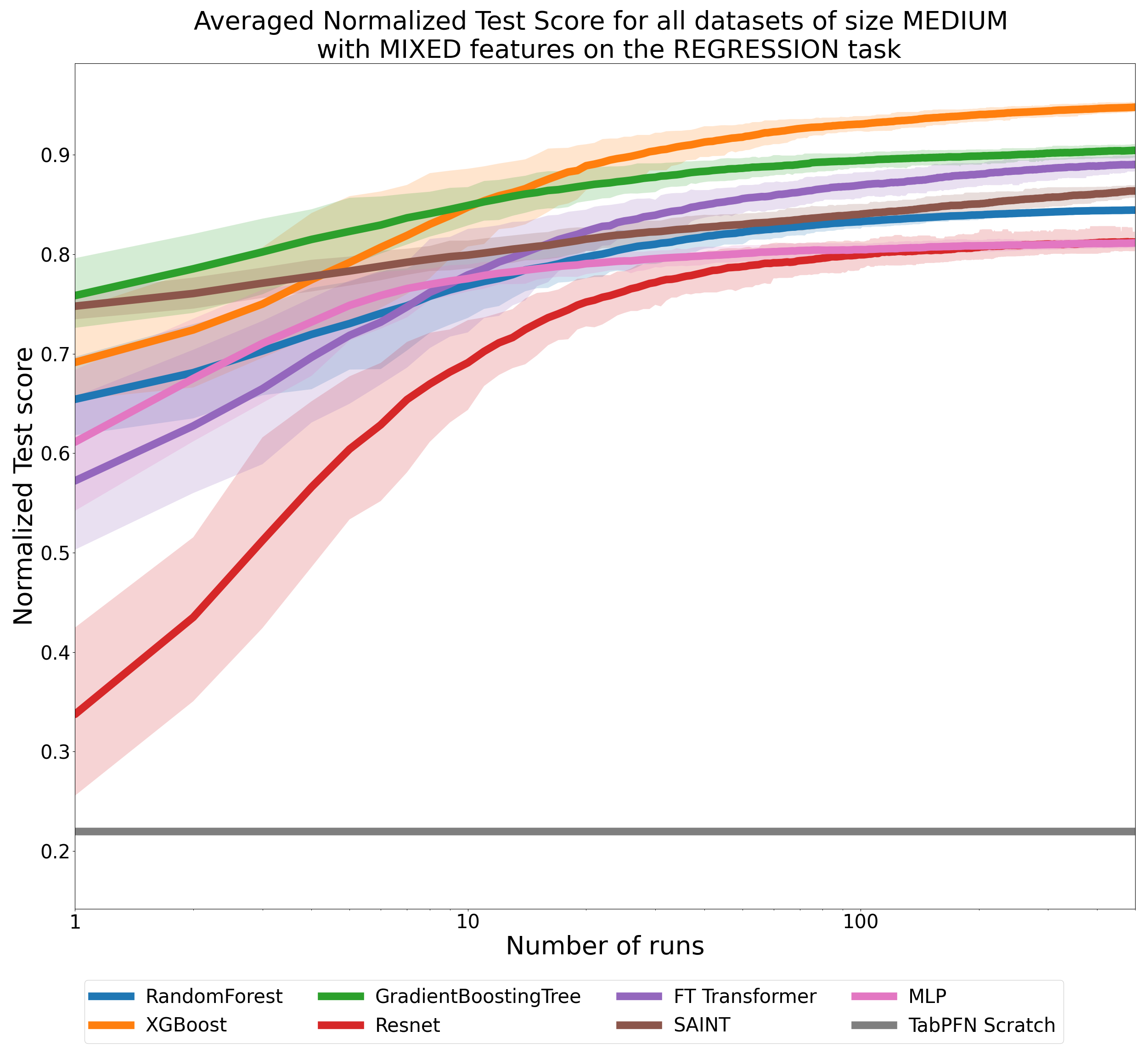}%
\label{fig:mixed_regression_summary}
\end{figure}

\clearpage
\section{Dataset Graphs} \label{sec:appendix_by_dataset}

\begin{figure}[hbt]
\centering
\includegraphics[width=0.65\textwidth]{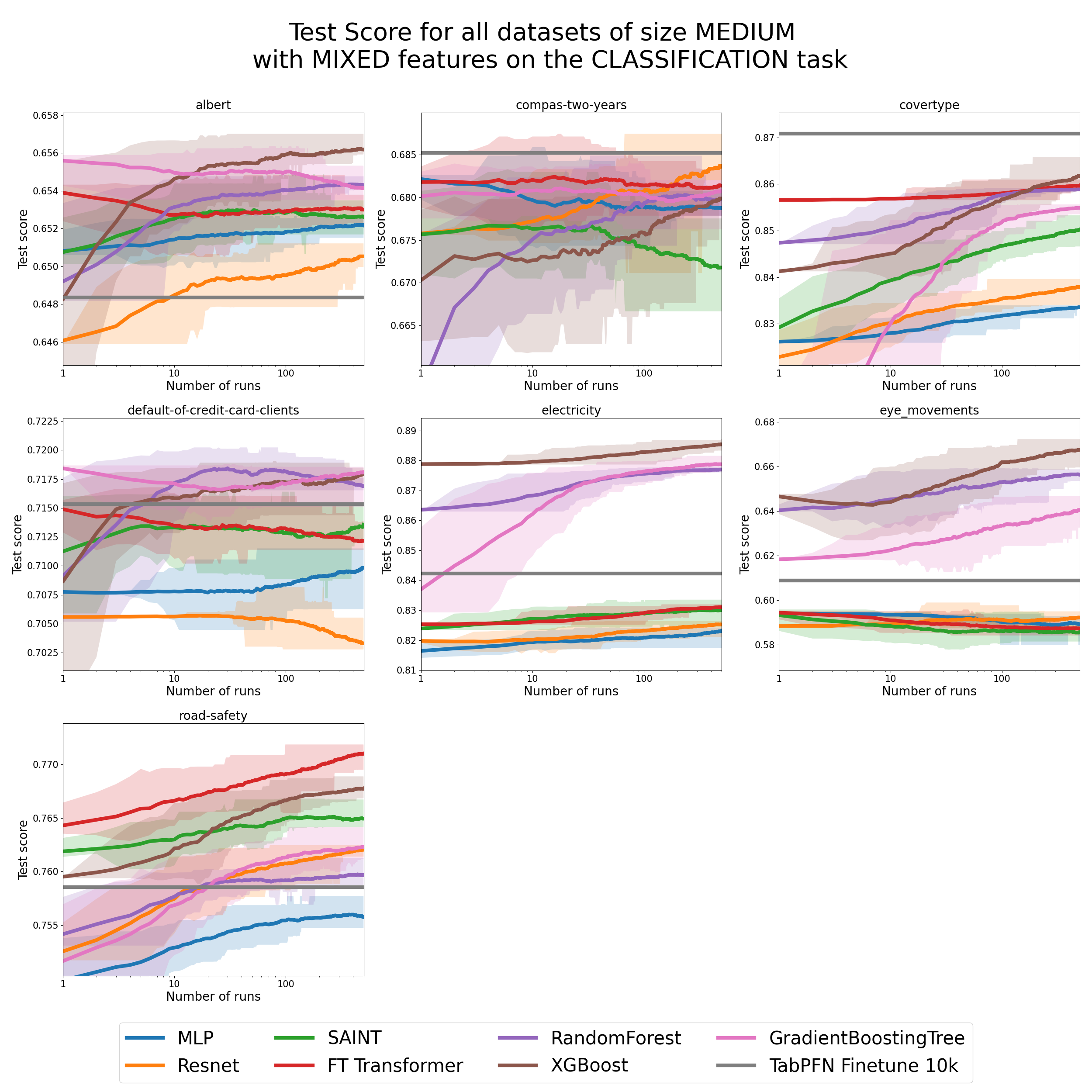}%
\label{fig:mixed_classification_datasets}
\end{figure}

\begin{figure}[hbt]
\centering
\includegraphics[width=0.65\textwidth]{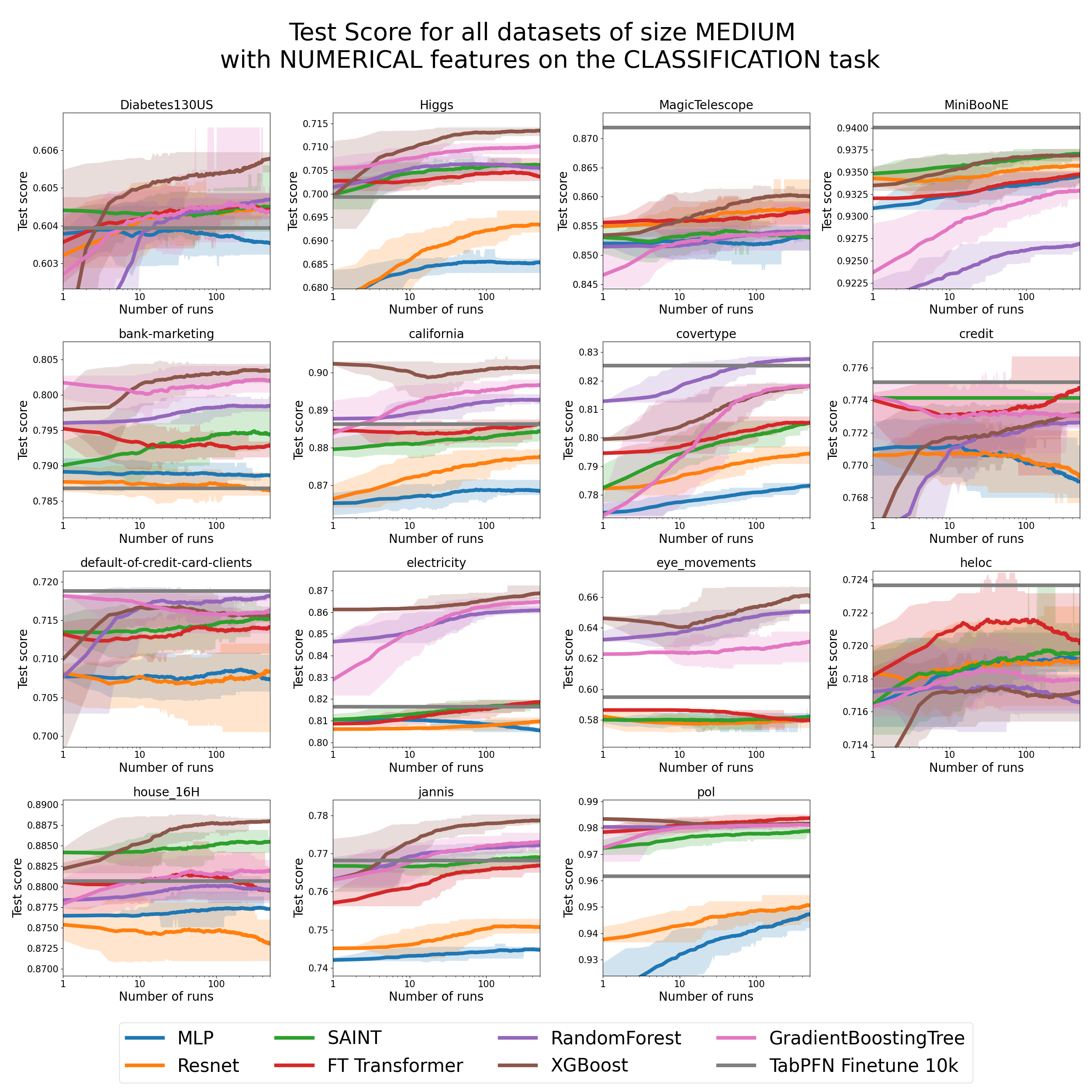}%
\label{fig:numerical_classification_datasets}
\end{figure}

\begin{figure}[hbt]
\centering
\includegraphics[width=0.65\textwidth]{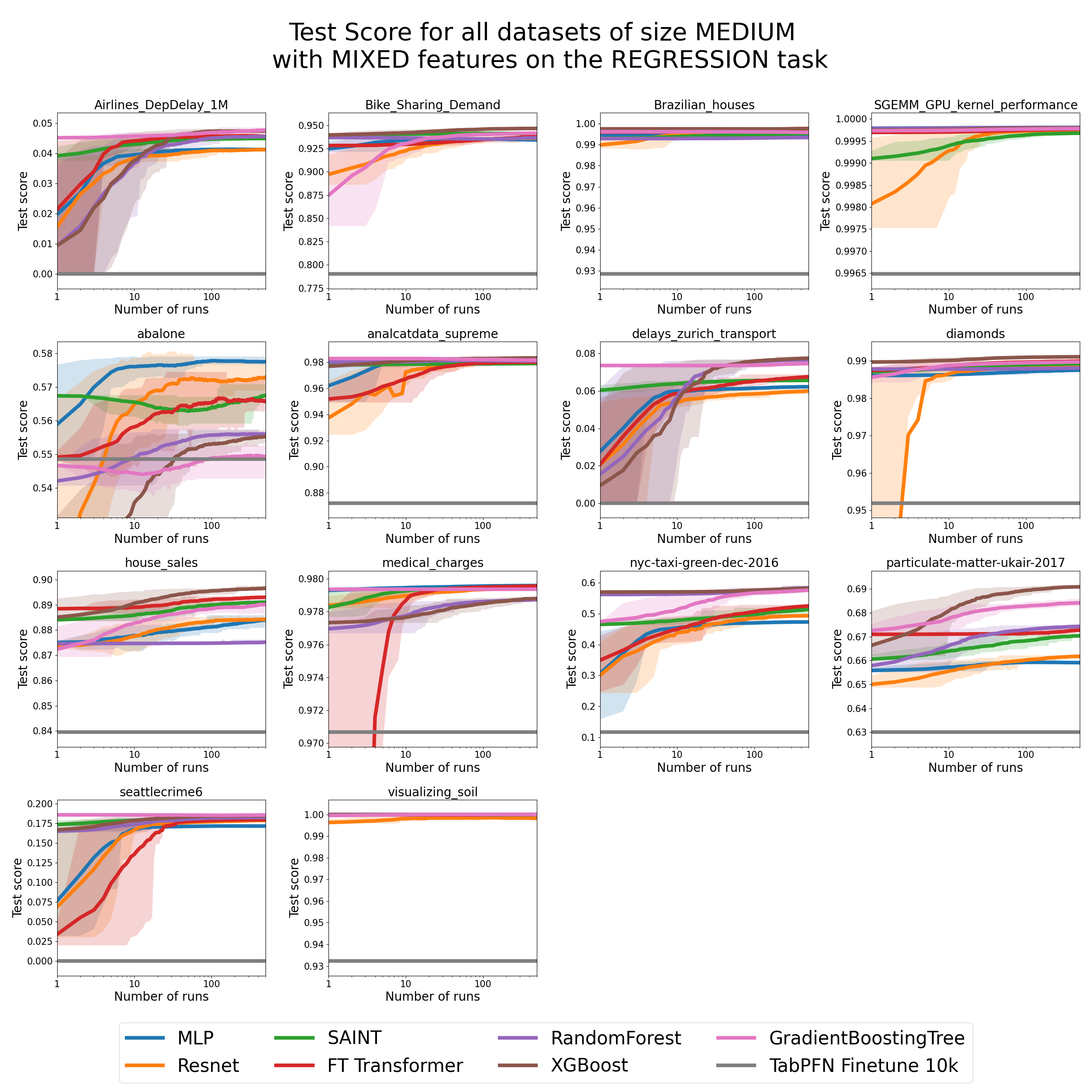}%
\label{fig:mixed_regression_datasets}
\end{figure}

\begin{figure}[hbt]
\centering
\includegraphics[width=0.65\textwidth]{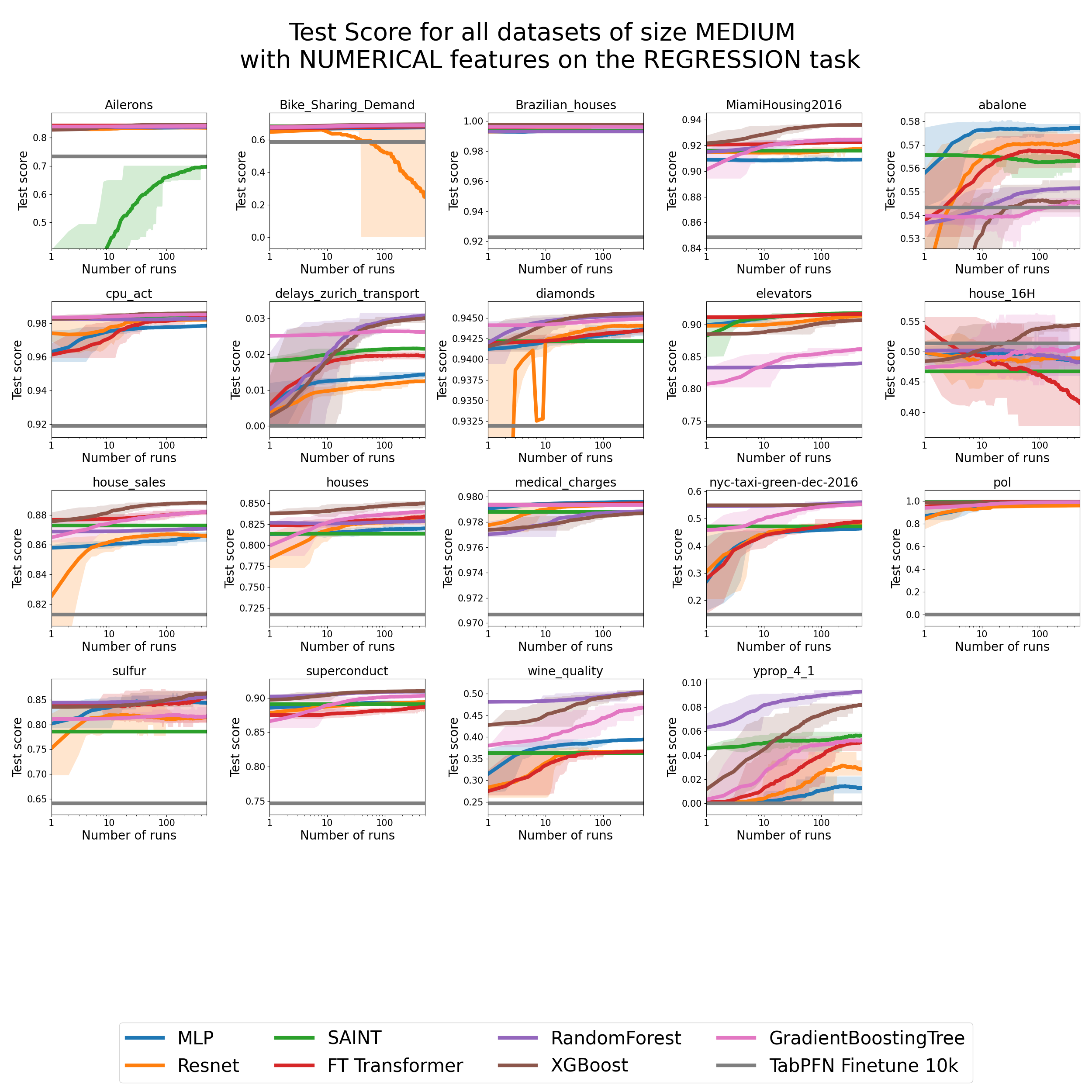}%
\label{fig:numerical_regression_datasets}
\end{figure}